%% file: acl_latex.tex
\pdfoutput=1

\documentclass[11pt]{article}

\usepackage[]{acl}

\usepackage{times}
\usepackage{latexsym}

\usepackage[T1]{fontenc}

\usepackage[utf8]{inputenc}

\usepackage{microtype}

\usepackage{inconsolata}

\usepackage{hyperref}
\usepackage{url}

\usepackage{subcaption}
\usepackage{multirow}
\usepackage[shortlabels]{enumitem}
\usepackage{microtype}
\usepackage{graphicx}
\usepackage{booktabs}
\usepackage{hyperref}
\usepackage{ifthen}
\usepackage{cuted}
\usepackage{makecell}
\usepackage{tabularx}
\usepackage{multicol}
\usepackage{wrapfig}

\usepackage{amsmath}
\usepackage{amssymb}
\usepackage{amsthm}
\usepackage{dsfont}
\usepackage{multirow}
\usepackage[shortlabels]{enumitem}
\usepackage{setspace}

\usepackage{bbm}
\usepackage{hyperref}

\input{math.tex}

\usepackage[ruled,boxed,lined]{algorithm2e} 
\let\oldnl\nl
\newcommand{\nonl}{\renewcommand{\nl}{\let\nl\oldnl}}

\title{Reward Gaming in Conditional Text Generation}

\author{Richard Yuanzhe Pang$^1$~~~~Vishakh Padmakumar$^1$~~~~Thibault Sellam$^2$ \\ 
{\bf Ankur P. Parikh$^2$~~~~He He$^{1}$}\\
$^1$New York University\\
$^2$Google DeepMind \\
{\tt yzpang@nyu.edu}}

\begin{document}
\maketitle
\begin{abstract}
To align conditional text generation model outputs with desired behaviors, there has been an increasing focus on training the model using reinforcement learning (RL) with reward functions learned from human annotations. Under this framework, we identify three common cases where high rewards are incorrectly assigned to undesirable patterns: noise-induced spurious correlation, naturally occurring spurious correlation, and covariate shift. We show that even though learned metrics achieve high performance on the distribution of the data used to train the reward function, the undesirable patterns may be \textit{amplified} during RL training of the text generation model. While there has been discussion about reward gaming in the RL or safety community, in this discussion piece, we would like to highlight reward gaming in the natural language generation (NLG) community using concrete conditional text generation examples and discuss potential fixes and areas for future work. 
\end{abstract}

\input{sec_1_intro}

\input{sec_2_3_background}

\input{sec_4-6}

\section*{Acknowledgement}

We thank Nitish Joshi, Nicholas Lourie, Angelica Chen, Chen Zhao, Sebastian Gehrmann, and anonymous reviewers for valuable discussion and feedback. This work is supported by the Google Research Collabs program. In addition, VP and HH are supported by Samsung Advanced Institute of Technology (under the project \textit{Next Generation Deep Learning: From Pattern Recognition to AI}) and AWS AI.

\bibliography{anthology,custom}


\appendix
\input{appendix}

\end{document}

%% file: math.tex
\usepackage{amsmath,amsfonts,bm}

\def\eqref#1{equation~\ref{#1}}

\def\1{\bm{1}}

\def\vx{{\bm{x}}}
\def\vy{{\bm{y}}}

\DeclareMathAlphabet{\mathsfit}{\encodingdefault}{\sfdefault}{m}{sl}
\SetMathAlphabet{\mathsfit}{bold}{\encodingdefault}{\sfdefault}{bx}{n}



%% file: sec_1_intro.tex
\section{Introduction}
\label{sec:intro}


Natural language generation aims to automatically produce text that is fluent, relevant, and factual. To train text generators such that the outputs are aligned with desired behaviors, recent work has used rewards learned from human annotations, such as improving the quality of generated summaries by using learned saliency and faithfulness metrics  \citep{pasunuru-bansal-2018-multi} and by using rewards based on learned question answering systems \citep{gunasekara-etal-2021-using-question}; the recent ChatGPT model also uses an approach in the same class. In general, this class of methods (1) collects a human annotation dataset $\mathcal{D}_{\text{reward}}$ consisting of, e.g., direct ratings of generations \citep{sellam-etal-2020-learning,nakatani-etal-2022-comparing,ramamurthy2022reinforcement}, labels of error spans in the generations \citep{freitag-etal-2020-human,amrhein-sennrich-2022-identifying}, or pairwise comparison of generations  given the same source sequence \citep{stiennon2020learning,wu2021recursively,bai2022training}; 
(2) learns a proxy reward function that scores generations (as opposed to a true reward function which is often given by human judgment) on $\mathcal{D}_{\text{reward}}$; and then (3) learns the text generator on a dataset $\mathcal{D}_{\text{task}}$, using RL with the learned reward function. 

What could go wrong when we obtain the reward signal from humans? The rewards would rarely be robust. 
When training the text generator, the distribution induced by the policy (i.e., the generator) changes because we frequently update it, which opens up opportunities for exploiting errors in the reward. Thus, even if the reward function performs well on the dev/test split of $\mathcal{D}_{\text{reward}}$ as an evaluator, the reward can still be gamed during RL training of the generator. Reward gaming commonly refers to the issue that when the proxy reward increases, the true reward decreases or stays stable \citep{amodei2016concrete,skalse2022defining}. 
In this discussion and in the context of NLP, we use ``reward gaming'' to broadly refer to the phenomenon that as training progresses, models produce low-quality generations that exhibit undesirable patterns while converging to high rewards.

Reward gaming can happen when an undesirable pattern is associated with a high reward in the learned metric. We identify three ways this phenomenon can happen. (1) A group of examples is misannotated systematically. For instance, suppose we train a model to do effective negotiation and annotators carelessly label all long paragraphs as effective, then the reward model would assign high scores on long generations even if they are nonsensical, and the generator would subsequently exploit this pattern. 
(2) $\mathcal{D}_{\text{reward}}$ contains some bias due to the data we select to annotate, or due to the people we select to be annotators. An example in the former case is that suppose every translation that contains ``\texttt{united nations}'' happens to have high quality/reward, possibly due to the way we collect $\mathcal{D}_{\text{reward}}$; then the neural machine translation model may end up almost always generating the {phrase} surrounded by some gibberish. An example in the latter case is that due to the selection bias of annotators, certain language varieties may be rated higher (or lower) by annotators, even if the language variety itself is not an indicator of quality \citep{plank2016non,sap-etal-2019-risk}; subsequently, the generator could learn to favor generating sentences of certain language varieties over others. (3) $\mathcal{D}_{\text{reward}}$ does not cover certain groups of sentences. A quick example is that a dialogue agent trained to negotiate generates incomprehensible sentences, because those sentences are underspecified by the reward function \citep{lewis-etal-2017-deal}. 

In short, among these three cases, the first two cases induce spurious correlations between the undesirable pattern and the reward, and the third case induces underspecified behavior on uncovered examples.

We use synthetic and real-world examples to illustrate the above three cases: even if the learned reward achieves a good performance on $\mathcal{D}_{\text{reward}}$, high rewards can still be assigned to undesirable patterns. Notably, we show that these patterns get \textbf{amplified} during RL training of the generators. 
For instance, a synthetic experiment discussed later (\S\ref{sec:noise-spurious}) shows it is possible that even a reward function that gives the correct reward on 99.3\% of the test split of $\mathcal{D}_{\text{reward}}$ can lead to generation failure after RL. 

We also review potential fixes (\S\ref{sec:remedies}), including restricting the policy -- e.g., maximum likelihood regularization which is commonly used in recent work including \citet{stiennon2020learning} and \citet{ramamurthy2022reinforcement} -- and fixing the reward itself like iteratively collecting human annotations. In light of these observations, we would like to bring more attention to reward gaming in the context of conditional text generation. Leveraging learned metrics during RL is a promising approach to training aligned text generation systems. But given that the rewards can only reliably improve generators if the sampled texts are within the distribution of $\mathcal{D}_{\text{reward}}$, \emph{extra caution is needed when interpreting the results when training text generators using learned rewards -- quality control or manual inspection is required to ensure good generation quality.}

%% file: sec_2_3_background.tex
\section{Related Work}
\label{sec:related}

Reward gaming or similar ideas have been {discussed} since \citet{goodhart1975problems}.
More recently, it is extensively discussed in \citet{amodei2016concrete}. In this discussion, we avoid the term ``reward hacking'' because reward tampering \citep{everitt2021reward} -- actively changing the reward (e.g., by execution of reward-modifying code under certain circumstances in a video game)
-- is also reward hacking, but it is not the topic of our discussion. 

Many examples have demonstrated the reward gaming behavior, usually in gameplay or autonomous driving. For example, in a boat racing game in \citeauthor{amodei2016concrete}, the boat would hit objects in circles mid-way in the race instead of completing the race (the latter being the intended goal), because the reward increases faster by hitting a certain set of objects than completing the race; 
\citet{baker2020emergent} find that the reward is gamed in a hide-and-seek game -- one behavior is that hiders can trap themselves using walls and boxes so the seeker never reaches them; 
the reward can be gamed in a tic-tac-toe game by making specific moves to cause opponents' out-of-memory crash and lead them to forfeit \citep{lehman2020surprising}. Similar reward gaming behaviors have been observed in Atari games \citep{ibarz2018reward,toromanoff2019deep}, in code/program generation \citep{lehman2020surprising}, in a football simulator \citep{kurach2020google}, in a neuromusculoskeletal environment where an agent learns to run \citep{kidzinski2018learning}, and so on. 

Reward gaming is rarely concretely discussed in conditional text generation. A quick example by \citet{lewis-etal-2017-deal} and \citet{kenton2021alignment} is that a dialogue agent trained to do successful negotiation ends up generating nonsensical sentences, because those generations are underspecified by the reward function that is used to train the dialogue model. 

Recently, there have been two findings that indicate the seriousness of reward gaming, albeit not in the context of NLP. First, more capable models may exacerbate reward gaming: \citet{pan2022the} study the reward gaming problem using traffic control, COVID response, blood glucose monitoring, and the River Raid game, by designing misaligned proxy reward functions; they find that if an agent is more capable (depending on, e.g., model size, the number of training steps), then it is better at exploiting loopholes in the reward function, and therefore ends up with a lower true reward compared to a less capable model. 

More recently, \citet{skalse2022defining} has suggested a strict definition of the hackability of a pair of reward functions, where ``a pair'' can be understood as an original reward and a proxy reward.\footnote{In short, reward functions $r_1$, $r_2$ are hackable w.r.t. a policy set and an environment, if there exist policies $\pi, \pi'$ such that $J_1(\pi) < J_1(\pi')$ but $J_2(\pi) > J_2(\pi')$ where $J_i$ denotes the expected return corresponding to reward function $r_i$. See Definition 1 in \citet{skalse2022defining} for details.} They find that the pair of non-trivial unhackable reward functions does not exist theoretically. The question then becomes whether it is safe to use a proxy reward function empirically. 

In this discussion, we aim to demonstrate the effect of reward gaming in text generation using concrete examples. Here are the two main differences of our discussion from the aforementioned examples. First, we focus on conditional text generation; in particular, the experiments in this discussion do not rely on state-of-the-art large language models -- we aim to use smaller specialized conditional generation models instead. Second, we aim to investigate the reward gaming categories when the reward signal is learned from human annotations.

\section{Background}
\label{sec:background}

Conditional text generation systems usually model $p(\vy \mid \vx)$ where $\vx=(x_1, \dots, x_{T_s})$ is a source sequence and $\vy = (y_1, \dots, y_T)$ is a target sequence. Most models use an autoregressive factorization: $\log p(\vy \mid \vx) = \sum_{t=1}^T \log p_\theta(y_t \mid \vy_{<t}, \vx)$, where $\vy_{<t} = (y_1, \dots, y_{t-1})$, and $p_\theta$ is parameterized with a neural network. Maximum likelihood estimation (MLE) leads to mismatched train/test history and objectives during sequence generation \citep{bengio2015scheduled,huszar2015not,ranzato2016sequence,schmidt-2019-generalization,pang2021text,arora-etal-2022-exposure}. In addition, recent work aims to better align training objectives with human-annotated quality of generated texts (e.g., translation quality judgments, summarization faithfulness, human preference of generations). 

The generation process can be considered a sequential decision making process suitable for RL. 
Given state $s_t=(\vx, \vy_{<t})$, the policy $\pi_\theta$ (i.e., $p_\theta$) takes action $a_t$ (a token in the vocabulary), transits to the next state $s_{t+1}$, and receives a reward $r_t \in \mathbb{R}$ 
learned from human annotations. 
Assume discount factor $\gamma=1$. To maximize the objective $J(\theta) = \mathbb{E}_{\tau\sim \pi_\theta} R(\vx, \vy)$, where $R(\vx, \vy) = { \sum_{t=1}^T  r_t }$, one way is to use policy gradient \citep[REINFORCE;][]{williams1992simple,sutton1999policy}: $\nabla_\theta J(\theta) = \mathbb{E}_{\tau\sim \pi_\theta} \sum_t \nabla_\theta \log \pi_\theta(a_t\mid s_t) \hat{Q}(s_t, a_t)$, where $\hat{Q}(s_t, a_t) = \sum_{t'=t}^T r_{t'}$
is the estimated return. Our work uses REINFORCE with tricks of advantage estimation and value function fitting, described in the appendix. Recently, proximal policy optimization \citep[PPO;][]{schulman2017proximal} has also been widely used. It aims to avoid reward performance collapse, but we argue that the choice of algorithm that makes generations achieve high rewards is orthogonal to the issue that high rewards can correspond to undesirable generations.

To stabilize RL training, in each RL training run, we first initialize the model using an MLE-trained model to ensure a good starting point for RL optimization. In addition, we also use KL regularization which helps RL optimization \citep{jaques2019way, stiennon2020learning,ramamurthy2022reinforcement}, so $J(\theta) = \mathbb{E}_{\tau\sim \pi_\theta} [ R(\vx, \vy) - \beta [\log \pi_\theta (\vy \mid \vx) - \log p_{\text{MLE}} (\vy \mid \vx)]]$ where $p_{\text{MLE}}$ is the model trained using standard MLE. To demonstrate reward gaming behaviors, we tune $\beta$ to achieve the highest validation reward in the synthetic Sudoku experiments, unless explicitly mentioned. Larger $\beta$, but not too large, likely leads to higher true reward \citep{gao2022scaling}, but $\beta$ is hard-to-tune. But in some examples (e.g., \S\ref{sec:covariate}), even large $\beta$ does not eliminate undesirable behaviors. We will discuss using KL regularization as a remedy in \S\ref{sec:remedies}.

%% file: sec_4-6.tex
\section{Examples of Reward Gaming in Conditional Text Generation}
\label{sec:gaming}

As a reminder, we consider the class of conditional text generation learning algorithms where we:
\begin{enumerate}[{(1)}]
    \item have a human annotation dataset $\mathcal{D}_{\text{reward}}$; 
    \item use this dataset to train a reward function $f_\phi$ that scores generations;
    \item learn the text generator on a dataset $\mathcal{D}_{\text{task}}$, using RL with the learned reward function. 
\end{enumerate}

Reward gaming happens when some undesirable pattern is associated with a high reward. We identify three such scenarios:
\begin{enumerate}[{(1)}]
    \item spurious correlation due to annotation errors;
    \item naturally occurring spurious correlation;
    \item underspecified behavior in the reward function due to covariate shift. 
\end{enumerate}

We use both synthetic and real-world tasks to demonstrate the reward gaming behavior. The full experimental details can be found in the appendix. The experiments in this discussion do not rely on large language models; instead, we aim to build smaller specialized conditional generation models. 

For synthetic tasks, we simulate all three settings using the following framework. We adapt Sudoku as a conditional text generation task.\footnote{Controlling spurious correlations in the reward is difficult on experiments using real-world generation tasks. Therefore, we rely on the Sudoku framework, which has all the key elements we need for such experiments: (1) it is a conditional generation task (where the model needs to learn the relation between the input and the output); (2) it has clearly defined ground-truth rewards which enable easy evaluation; (3) it allows for easy manipulation of spurious correlations in the reward function. Therefore, we use the Sudoku experiments to show that reward gaming exists in conditional generation, and the reward gaming effect can be severe.} 
A valid Sudoku is a 9x9 grid with each cell containing a number from 1 to 9, such that no rows/columns and none of each of the nine non-overlapping 3x3 regions contains duplicates. For this task, let the input be the first $k$ ($k$ randomly chosen from 36 to 80) cells in a valid Sudoku after flattening it row by row. Let the reference output be the rest of the cells (i.e., the last $81-k$ cells). The goal is to generate the continuation to form a \textit{valid} Sudoku, given the prefix (i.e., first $k$ cells). To measure generation quality, we define \textit{success rate} to be the percentage of generations that result in valid Sudokus.

While the sequence generator can be rule-based without using neural nets in this synthetic setting, to illustrate reward gaming, we consider learning the generator from a learned reward function. 

\subsection{Noise-Induced Spurious Patterns}
\label{sec:noise-spurious}

We want to study settings where there is noise in human annotations. If we inject a small amount of high-reward but low-quality examples in $\mathcal{D}_{\text{reward}}$, the reward function could put a high reward incorrectly on these examples. 

\paragraph{Synthetic example: modified Sudoku.}

$\mathcal{D}_{\text{reward}}$ is a balanced dataset containing 500k positive and 500k negative examples. Out of the 500k positive examples, 0.5k (0.05\% of all examples) are false positives, i.e., invalid Sudokus. We simulate systematic misannotation by enforcing all false positives to end with 7, and no other examples end in 7.\footnote{For positive examples, we first create a set of 2M valid Sudokus, and then sample from the set. Many negative examples are small modifications of positive examples (\S\ref{app:noise}) to ensure a high-quality $f_\phi$.} This design is intended to simulate systematic errors in human annotation; e.g., a group of sentences on rare topics getting mislabeled.

The reward is the probability of the Sudoku being valid, estimated by a classifier $f_\phi$. 
$f_\phi$, based on a tiny RoBERTa (\S\ref{app:noise}), achieves 99.3\% accuracy on the \textit{i.i.d.} test split of $\mathcal{D}_{\text{reward}}$. 
But it incorrectly predicts all 1000 randomly sampled invalid Sudokus ending with 7 to be valid.\footnote{The reward makes the wrong prediction on those examples, but they represent a small portion of the dataset used to train the reward.}

\begin{figure}[t]
     \centering
     \begin{subfigure}[t]{0.47\columnwidth}
         \centering
         \includegraphics[width=\columnwidth]{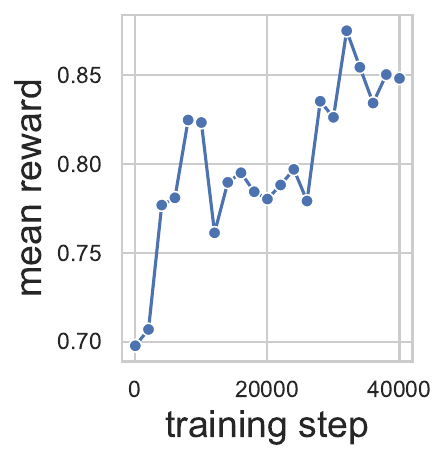}
         \label{fig:sdk-reward}
     \end{subfigure}
     \begin{subfigure}[t]{0.455\columnwidth}
         \centering
         \includegraphics[width=\columnwidth]{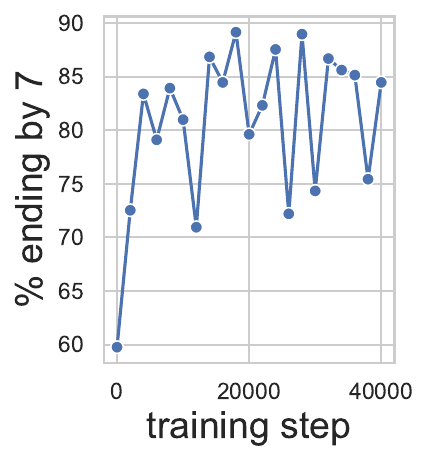}
         \label{fig:sdk-ending}
     \end{subfigure}
    \vspace{-7mm}
    \caption{Left: mean reward vs. training step. Right: mean \% of sampled sequences that end with 7 vs. training step. Each point corresponds to the mean value for a bucket of 2,000 training steps. Soon after training starts, the vast majority of sequences would end with 7; the \% of valid continuations is always $<$15\%. {Another training run using a variant of the training algorithm shows similar trends (see \S\ref{app:background} and \S\ref{app:noise}).}
    }
    \label{fig:sdk}
\end{figure}

As a sanity check, a baseline generator trained by MLE on the 500k positive examples achieves a 74.7\% success rate in spite of the noise. 
However, the RL-trained generator produces a large fraction of invalid generations that end in 7 despite achieving a high reward. 
Figure~\ref{fig:sdk} shows that the reward increases to above 0.8 (a large reward given the range $[0,1]$),  and the amount of Sudokus ending with 7 oscillates around 85\%; however, only 0.1\% of the actual correct reference generations end with 7. Additionally, given a reward of 0.85 in the figure, we would expect around 85\% of generations to be valid; however, the success rate (i.e., the proportion of valid generations) turns out to be always smaller than 15\% throughout training. 

In short, in this specific example, even 0.05\% of noise in $\mathcal{D}_{\text{reward}}$ could lead to generation failure ($>$80\% of generations are invalid), as the RL training of the generation model amplifies the failure mode.

\paragraph{Experimental details for the above example.}

The RoBERTa-tiny-based \citep{liu2019roberta} reward function has 4 encoder layers and 2 attention heads; the encoder embedding dimension is 64, and the dimension for FFN for 256. 
For the sequence generator,  we use a smaller version of the \texttt{transformer\_iwslt\_de\_en} architecture in \texttt{fairseq} \citep{ott-etal-2019-fairseq}. The encoder embedding dimension and the decoder embedding dimension are both 32. We use 2 attention heads in both the encoder and the decoder. The dimension for FFN in both the encoder and the decoder is 64. There are 2 encoder layers and 2 decoder layers. 
Please refer to the appendix for more details.

\paragraph{Takeaway.} Even a small amount of noise in $\mathcal{D}_{\text{reward}}$ can enable the reward function to assign high reward on sequences containing certain undesirable patterns. After RL training, a large proportion of generations could incorrectly contain those undesirable patterns.

\subsection{Naturally Occurring Spurious Patterns}

The spurious correlation is not necessarily noise-induced but can be naturally occurring. Due to the selection bias of annotators, certain language varieties may be preferred over others \citep{plank2016non,sap-etal-2019-risk,korbak2022rl}, although language varieties do not indicate quality in many tasks. In addition, due to the selection bias of examples that are annotated, some attributes that are irrelevant to the quality get correlated with the reward \citep{wiegreffe2021teach,pezeshkpour-etal-2022-combining}. 
If high rewards are assigned to these spurious patterns (e.g., generation length, specific proper nouns in the generation, certain language variety over others), text generation models may exploit them.  

\begin{table}[ht!]
\setlength{\tabcolsep}{2.1pt}
\centering
\begin{tabular}{lcc}
    \toprule
    & correct & incorrect \\ 
    \midrule
    repeat & 0 (n/a) & 13,053 (0.670) \\
    no repeat & 9,638 (0.999) & 123,645 (0.983) \\
\bottomrule
\end{tabular}
\caption{Contingency table for the first 1500 training steps. Correct: the generation is valid; repeat: there is repetition in the last nine numbers of the output. Inside the parentheses: average reward. Most continuations are unrepetitive; they have high rewards but most (92.8\%) are incorrect. 
}
\label{tab:sdk-2}
\end{table}

\paragraph{Synthetic example: Sudoku revisited.}

$\mathcal{D}_{\text{reward}}$ is dataset with 200k randomly sampled valid Sudokus as positive examples and 200k randomly sampled invalid Sudokus as negative examples. Using this dataset, we simulate the setting where a simple feature (the feature that ``the last nine numbers of the output do not repeat'') is predictive of the reward (validity) on a biased $\mathcal{D}_\text{reward}$. Repetitions co-occur with 99.9\% of negative examples, and therefore the repetition is a highly predictive feature of the reward. 

The reward function, $f_\phi$, achieves 99.9\% accuracy on the test split of $\mathcal{D}_{\text{reward}}$. We then train the conditional text generation model using RL where $f_\phi$ is the reward.

Table~\ref{tab:sdk-2} shows that when training the text generator, the model exploits the non-repetition pattern that leads to high reward, but the vast majority of such sequences (92.8\%) are in fact incorrect.

\paragraph{Real-world example: machine translation (MT) using dense reward.}

\begin{figure}[t]
     \centering
     \begin{subfigure}[t]{0.327\columnwidth}
         \centering
         \includegraphics[width=\columnwidth]{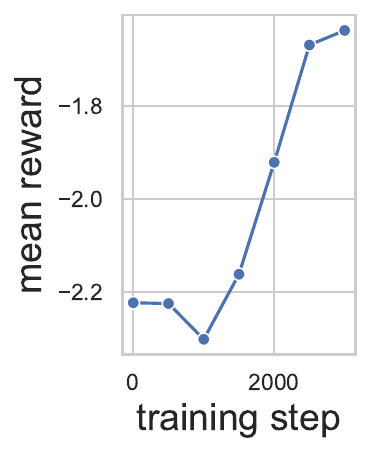}
         \label{fig:mqm-reward}
     \end{subfigure}
     \begin{subfigure}[t]{0.31\columnwidth}
         \centering
         \includegraphics[width=\columnwidth]{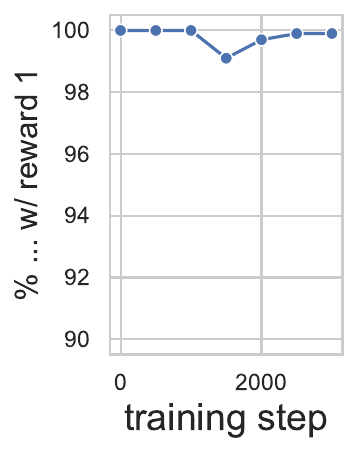}
         \label{fig:mqm-token-reward}
    \end{subfigure}
     \begin{subfigure}[t]{0.309\columnwidth}
         \centering
         \includegraphics[width=\columnwidth]{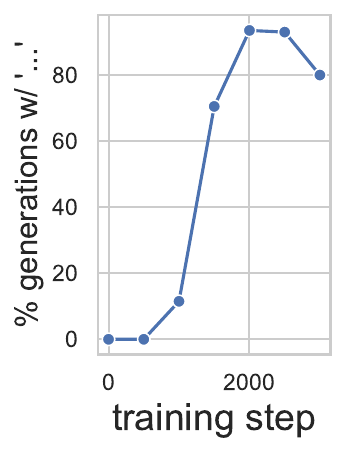}
         \label{fig:mqm-pattern}
     \end{subfigure}
    \vspace{-7mm}
    \caption{Left: mean sequence reward vs. training step. Middle: mean reward of ``...'' vs. training step. Right: mean \% of sampled sequences that contain ``...'' vs. training step. During training, total (seq-level) reward increases; reward for ``...'' is always close to one; \% of sampled generations that contain ``...'' increases to $>$3/4.}
    \label{fig:mqm}
\end{figure}

The WMT MQM dataset \citep{freitag-etal-2021-experts} is a high-quality human annotation dataset on translations, where each Zh-En translation is annotated with $\leq$ 5 most serious 
error spans by expert annotators according to the MQM metric \citep{lommel2014multidimensional}. Each of the $\leq$ 5 spans is annotated with no error, minor error, or major error. 
In $\mathcal{D}_{\text{reward}}$, an example annotation of a generated translation is as follows: 
\begin{quote}
    \small
    state-owned enterprises and <major> {\color{gray} advantageous} </major> private enterprises entered the <major> {\color{gray} revolutionary base area} </major> <major> {\color{gray} of} </major> <minor> {\color{gray} south ji@@ ang@@ xi} </minor> .
\end{quote} 
Major errors are between the ``major'' tags, and minor errors are between the ``minor'' tags. The source sentences of MQM annotations come from WMT Chinese to English (Zh-En) sets \texttt{newstest2020} and \texttt{newstest2021} \citep{mathur-etal-2020-results,barrault-etal-2020-findings,akhbardeh-etal-2021-findings}, as well as TED talks from WMT2021 \citep{freitag-etal-2021-results}. Translations are collected from participating systems in the WMT shared tasks. Human-written references are also integrated into the annotation dataset.

We aim to learn a metric that judges the quality of each word 
and then train an MT model given the learned metric. 
$f_\phi$ is a scorer that predicts whether each token in a given translation is in a no-error span. 
Let the reward $r_t$ be the score that $f_\phi$ outputs at time-step $t$. 
Our key observation is that certain tokens are spuriously correlated with no-error annotations in the dataset. The ellipses punctuation (``...'') is one of them: experts annotated 98.3\% of the occurrences as no-error. 

Figure~\ref{fig:mqm} shows that during RL training of the MT model on WMT17 Zh-En, as training goes on, the percentage of translations with ellipses increases and the ellipses achieve high rewards. The frequent occurrence of ellipses, however, are undesirable. Moreover, there is no improvement in BLEU: the MLE-trained translator achieves a dev BLEU of 23.1, but in different runs of our algorithm, the BLEU never exceeds 21.0. All experiments are run using a KL coefficient 0.1.\footnote{In comparison, \citet{bai2022training} use a coefficient of 0.001.}

In other training runs of $f_\phi$ and MT model, we found other tokens that are spuriously correlated with the reward. The token ``conduct'' is one example. 
Only 0.01\% of all tokens are ``conduct'' but about 95\% of them are labeled as ``no error.'' Here are some example model generations where ``conduct'' is integrated but its use is incorrect and nonsensical.

\begin{quote}
\small

\textit{Reference}: 66 countries and regions have reported the evidence of the mosquito-borne zika virus transmission since 2015.\\
\textit{Generation}: the 66 countries and regions have been able to \underline{conduct} the evidence in the dissemination of the virus in 2015.

\smallskip

\textit{Reference}: the ankara newspaper daily hurryiet, citing military sources, added turkey's military attaché in jordan was also missing.\\
\textit{Generation}: the newspaper in ankara has been able to \underline{conduct} the military information and the military work in jordan and the disappearance of military work.

\smallskip

\textit{Reference}: i found out after the sports day some of the parents went on facebook and made a fuss about it and that got a lot of people interested.\\
\textit{Generation}: the some parents have been able to \underline{conduct} the campaign day and the some comments on this matter and the many persons have been able to \underline{conduct} attention.

\end{quote}

In another run, we also find that ``in accordance with'' starts occurring often in generations. We hypothesize that even if we suppress particular tokens from occurring, the model could 
frequently generate other tokens or discover more spurious features that are more obscure and hard-to-interpret.

\paragraph{Experimental details on the above examples.}

For the Sudoku experiment, the hyperparameters are selected from the same sets as in \S\ref{sec:noise-spurious}. For the MT experiment, to train the classifier $f_\phi$, the model is initialized by a WMT17 Zh-En MLE-trained model. Then, the source sentence is fed into the encoder, and the target sentence is fed into the decoder. However, we remove the attention mask in the decoder that prevents hidden states at token $t$ from seeing future hidden states. The reward $r_t$ is the probability that the $t$-th token is erroneous, according to $f_\phi$. 
For $D_{\text{task}}$, our translation task uses the WMT17 Zh-En dataset, and $f_\phi$ is fine-tuned from an MLE-trained MT checkpoint using the WMT17 Zh-En dataset. We use a transformer model with 6 encoder layers and 6 decoder layers. The number of attention heads is 8 in both the encoder and the decoder. The FFN embedding dimension is 2048 in both the encoder and the decoder.

\paragraph{Takeaway.} Even a small amount of examples with spurious patterns in $\mathcal{D}_{\text{reward}}$ can enable the reward function to assign high reward on sequences containing those patterns. After RL training, a large proportion of generations could incorrectly contain those patterns.

\subsection{Covariate Shift}
\label{sec:covariate}


During RL training, the policy (i.e., the generator) may sample examples out of the support of the reward model. Therefore, in these examples, the reward model's behavior is underspecified -- it may or may not assign high rewards to these low-quality examples.

\paragraph{Synthetic example: another Sudoku variant.}

$\mathcal{D}_{\text{reward}}$ contains 200k positive and 200k negative examples.\footnote{Negative examples are obtained by swapping two different tokens of a positive example 1--20 times.} 
We design $\mathcal{D}_{\text{reward}}$ in such a way that the model behavior would be undefined for certain inputs. All examples end with 1; continuations that end with 2--9 are not in the support on the data used to train the reward function $f_\phi$. 

$f_\phi$ achieves 96.5\% accuracy on the test split of $\mathcal{D}_{\text{reward}}$. We sample 1000 in-support (i.e., ending with 1) and 1000 out-of-support (i.e., ending with 2--9) invalid Sudokus. The model only misclassifies 1 out of 1000 example as valid on the in-support set; in contrast, 659 out of 1000 examples are misclassified as valid on the out-of-support set. 

During RL training of the conditional text generation model, the reward for sampled generations increases above 0.8. We expect the reward to imply that more than 80\% continuations are estimated to be valid by the reward; however, only $<$10\% of the continuations are actually valid. 

\paragraph{Real-world example 1: AgreeSum.} 
One simple example reproduces the multi-doc AgreeSum summarization \citep{pang-etal-2021-agreesum}. The input of the task is a cluster of articles, and the expected output is a summary that is faithful to \textit{every} article in the cluster.  
We consider $\mathcal{D}_{\text{reward}}$ that consists of faithfulness annotations on article-summary pairs provided by the AgreeSum paper. 
The reward function $f_\phi$ is a summary-article faithfulness classifier. $f_\phi$ achieves 79\% dev accuracy, which we use as the reward. 
However, the shortest summary in $\mathcal{D}_{\text{reward}}$ is 7-token-long, so the behavior of the reward for shorter summaries is underspecified. 
Training a summarizer using the faithfulness classifier as the reward leads to short summaries -- most of which ($>$90\%) are $\leq$ 2 tokens. Even though these near-empty summaries can be technically considered as being entailed in the article, we have not specified in $\mathcal{D}_{\text{reward}}$ that these summaries are acceptable.

\paragraph{Real-world example 2: MT using BLEURT.}

\begin{figure}[t]
     \centering
     \begin{subfigure}[t]{0.47\columnwidth}
         \centering
         \includegraphics[width=\columnwidth]{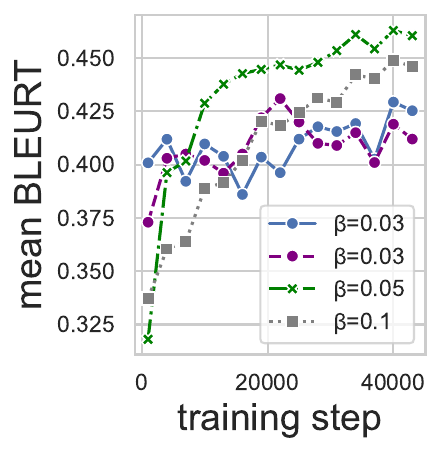}
         \label{fig:mt-bleurt}
     \end{subfigure}
     \begin{subfigure}[t]{0.47\columnwidth}
         \centering
         \includegraphics[width=\columnwidth]{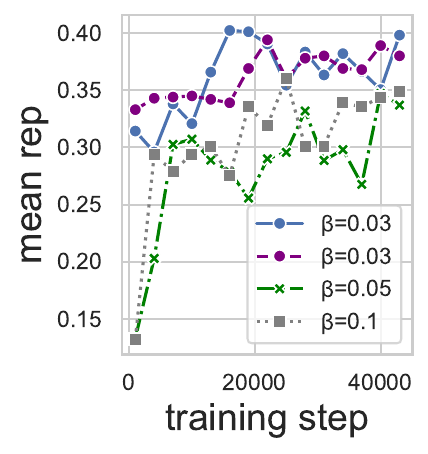}
         \label{fig:mt-rep}
     \end{subfigure}
    \vspace{-7mm}
    \caption{Left: mean BLEURT vs. training step. Right: mean $\mathrm{rep}$ vs. training step. Each point corresponds to the mean value for a bucket of 3,000 training steps. Each bucket contains $\geq$140 translations whose source sentences are longer than 180 tokens. We see that BLEURT increases during RL training; $\mathrm{rep}$ increases as well. $\mathrm{rep}$ for reference translations (whose source length $>$180) is 0.12, much smaller than achieved in our experiments. 93\% of translations has $\mathrm{rep}<$0.2. The two runs with $\beta=0.03$ use different baselines (see \S\ref{app:background}). Repetition is a problem even for large $\beta$.}
    \label{fig:mt}
\end{figure}

BLEURT \citep{sellam-etal-2020-learning} is a metric trained on expert annotations provided by WMT metric tasks. We train a text generator by RL using BLEURT-20-D3, a distilled version of BLEURT-20. BLEURT is trained on very few repetitive generations and very few long generations as discussed in the next paragraph. WMT15--19 human rating data \citep{stanojevic-etal-2015-results,bojar-etal-2016-results,bojar-etal-2017-results,ma-etal-2018-results,ma-etal-2019-results} are used to train BLEURT. 
We use BLEURT to train a MT model on the IWSLT14 De-En task \citep{cettolo2014report}. MLE-trained model achieves 63.9 in BLEURT on test set and RL-trained model achieves 65.5, so RL is successful judging by the increase in BLEURT. 

Repetitive translations are out-of-support in our case, where repetition ($\mathrm{rep}$) is measured the percentage of repeated 3-grams. In fact, only 0.02\% (58/247,157) translations have $\mathrm{rep}>$0.4 and 0.05\% translations have $\mathrm{rep}>$0.3 in $\mathcal{D}_{\text{reward}}$. In addition, long translations are also out-of-support: only 0.01\% of translations in $\mathcal{D}_{\text{reward}}$ has length longer than 180 BPE tokens.\footnote{BPE rules are learned using the IWSLT14 De-En dataset.} 

In the below analysis, we only examine the set of examples whose source length is larger than 180 tokens.\footnote{Longer source sequences likely imply longer translations. We threshold based on source length so that the BLEURT comparison later is fair.}
We find that BLEURT does not punish for excessive repetition in the samples during RL: average BLEURT for translations with $\mathrm{rep}>$0.4 ($>$40\% of 3-grams are repetitions -- an example is shown in the footnote to demonstrate that 40\% is an undesirably large proportion)\footnote{As an example, the following sentence has $\mathrm{rep}=0.397$: \texttt{pip was adopted from "great expectations; superman was a foster child; and the azbeth salander," the girl with the dragon tattoo, "was a foster child and a pure man; lyra belacqua from philip pullman," and a foster child, jane eyre, adopted, and roald's james, and the great, and he was a parent, and a parent, and then, "and then, you know," and then, "and then, you know," and then, "and, you know," the "-- and, you know," the "-- and, you know," the "the" -- and "you know," the "}} 
in the first 45,000 steps of training\footnote{Using $\beta=0.05$ which leads to the best dev BLEURT.} is 42.7, and average BLEURT for translations with $\mathrm{rep}<$0.2 is 42.3.\footnote{0.2 is an acceptable threshold, given that 93\% of translations whose source sentence length $>$180 have $\mathrm{rep}<$0.2.} So the reward does not discourage the MT model from generating repetitions. 

Next, we show in Figure~\ref{fig:mt} that as training goes on, translations get more and more repetitive as BLEURT increases. 
To summarize, given that long repetitive translations are rare in $\mathcal{D}_{\text{reward}}$, the reward is underspecified on them. This repetition pattern is not discouraged by the reward, and thus it is subsequently exploited by the MT model. 

\paragraph{Experimental details for the above examples.}

For AgreeSum, given URLs in the original dataset, to find the corresponding articles, we use the \texttt{newspaper3k} library. The reward function (classifier) is based on RoBERTa-large. The summarizer is based on BART-large \citep{lewis-etal-2020-bart}. For the MT experiment, the MT model has an embedding dimension of 512 for both the encoder and the decoder. The FFN embedding dimension is 1024 for both the encoder and the decoder. Both the encoder and the decoder have 4 attention heads and 6 layers. More details can be found in the appendix.

\paragraph{Takeaway.} Suppose examples with certain undesirable patterns occur only rarely (or never occur) in $\mathcal{D}_{\text{reward}}$. Then, they could be out of support of the reward model, and the reward model could assign high reward to these examples. After RL training, many generations could contain those patterns.

\section{Possible Remedies}
\label{sec:remedies}

As discussed in \S\ref{sec:related}, \citet{skalse2022defining} has suggested that a pair of unhackable nontrivial original-proxy reward functions do not exist in theory. Then, when is it safe to use the proxy reward function? While this is still an open question, it is possible to reduce the extent of generating undesirable sentences through the following approaches. 

The fundamental problem is that errors in the reward functions, specifically the over-confident errors where low-quality outputs have high rewards, can be exploited during RL training of text generators. Thus, one solution is to avoid OOD states that incur such errors by restricting the policy.

\paragraph{Restricting the policy by regularizing toward the ML solution.} A common strategy is to {regularize toward the ML solution}. In practice, we can interpolate RL and ML losses \citep{wu2016google}, interleave RL and ML updates \citep{lewis-etal-2017-deal,guo2018long}, or use KL-regularized RL \citep{jaques2019way, stiennon2020learning,ramamurthy2022reinforcement}. 
Here are a few potential issues. First, RL exploration could be important in case the reference dataset is small and consequently the ML solution is sub-optimal. In these tasks, it is often easier to verify or rate a generation than to provide a reference generation (unless we have access to a large language model). For example, in AgreeSum, there are not enough reference summaries due to data collection costs, but given a decent article-summary faithfulness classifier, we can discover new summaries that have high rewards. Similarly, in creative generation tasks like story generation and textual style transfer, or in code generation, there may not be a large enough high-quality reference dataset, but a reward function is often available. Second, ML solution may not be optimal even with an adequately large reference dataset; e.g., degeneracies like unreasonably short translations \citep{stahlberg-byrne-2019-nmt,kulikov2021characterizing} and repetitive generations \citep{welleck2020neural,welleck-etal-2020-consistency,chiang-chen-2021-relating} may often have high probabilities. 

By relying on ML, we are essentially optimizing toward a different objective from the reward \citep{korbak2022rl}; thus, we may need to find another automatic evaluator (instead of the proxy reward) to do hyperparameter tuning and model selection, which is difficult empirically and may require numerous tricks \citep{khalifa2021a}.\footnote{\citet{gao2022scaling} has recently discovered that larger coefficients for the KL penalty (for ML regularization) does not improve the frontier of the curve of the gold reward-model score vs. the KL divergence (between the RL-optimized model and the ML model), so the coefficients only impact gold reward's convergence speed. See Figure 9 of \citet{gao2022scaling}. Thus, tuning and empirical tricks would be crucial to the success of KL-regularized RL.} In addition, KL-regularized RL cannot enforce distributional conditions of the set of all generations \citep{khalifa2021a,santurkar2023whose}.


\paragraph{Restricting the policy by leveraging a discriminator.}

Following \citet{goodfellow2014generative}, \citet{pang-etal-2021-agreesum}, and \citet{vuong2022dual}, another idea similar to ML-regularization is to {leverage a discriminator} that distinguishes between sampled generations and the set of dataset-provided generations.\footnote{The discriminator predicts whether the generation is machine-generated or comes from \textit{the set of} references. This technique is useful when there are only few parallel datapoints.} During RL training, we force the model to produce generations that are indistinguishable from references according to the discriminator. Discriminator and RL updates are interleaved. It is difficult to use GAN to train a high-quality text generator, but we hypothesize that the discriminator can reduce easy-to-identify low-quality examples during RL training.

\paragraph{Fixing the reward itself.}

Another thread of remedies is to {fix the reward} itself. An effective approach is to \textbf{iteratively collect human annotations} \citep{stiennon2020learning,bai2022training,fan2022nano}: the reward is iteratively updated with human annotations on latest model generations; thus, the generations with low human preferences but high rewards will be corrected through this iterative process. 
One concern is the cost (e.g., may require bilingual speakers or professional translators for MT annotations), which may prohibit an adequate amount of iterations or adequately frequent iterations. \citet{krakovna2020specification} has discussed the possibility that a machine can learn to fool human evaluators in robotics, but it is unclear what the equivalence in conditional text generation is. 
So far, this approach has been successful, with the critical assumption that there is little budget/resource constraint to obtain enough high-quality annotations and enough iterations of annotations. 

Another caveat is that as \citet{perez2022discovering} has recently discovered, RL-with-human-feedback may amplify one-sided views (e.g., on political issues); they claim that this phenomenon can be explained by the selection bias of annotators, leading to unrepresentative reward. Similarly, if selection bias is unavoidable in the context of conditional text generation (therefore the unrepresentative reward is unavoidable), we may need another way of fixing the reward and preventing the generation model from amplifying the bias -- e.g., by \textbf{hard-coding a set of principles} as in Constitutional AI \citep{bai2022constitutional}. 

\paragraph{More discussion.}

An additional method in the RL literature is conservative Q learning \citep{kumar2020conservative}; it aims to push down all high rewards to ensure that the out-of-distribution states do not achieve high Q values, but the approach requires extensive hyperparameter tuning \citep{zheng2022online}. Another possibility to avoid the reward gaming issue is to simply avoid interaction with the environment using methods like \citet{pang2021text} to learn from demonstrations, so the errors in the reward function will be less exploited; additionally, non-RL objectives that can learn from both positive and negative examples \citep{adolphs2022cringe} are also potential solutions.

\section{Conclusion}

We use synthetic and real-world tasks to demonstrate that even if a learned reward achieves high performance on $\mathcal{D}_{\text{reward}}$, a high reward may still get assigned to undesirable patterns which get amplified during RL training of the conditional text generation model. A critical future direction is to detect obscure or hard-to-interpret gaming behaviors especially in long generations. Then, we can investigate when or how easily a spurious feature could be exploited, by exploring the relationship among the minimum description length of a spurious feature \citep{voita-titov-2020-information} or similar statistics, the proportion of datapoints that contains the spurious feature, the choice of RL algorithm, and the degree of the reward gaming behavior. 
Additionally, research on new approaches of reward or preference learning is needed.

\section*{Limitations}

First, off-policy algorithms like Q learning are not explored in this discussion. 
Second, the reward gaming issue is not a novel topic in the RL community for tasks like gameplay or autonomous driving \citep{amodei2016concrete,koch2022objective}. However, we hope to highlight issues in the NLG community (specifically on conditional text generation tasks without use of large language models) especially given the recent endeavors on learning from learned metrics. 

In addition, the paper aims to demonstrate the existence of reward gaming in conditional text generation, not the certainty regardless of experimental settings (hyperparameters, architectures, etc.). Given that our experiments use reasonable settings which lead to degenerate texts, we argue that reward gaming could be a common issue when learning a text generation model using RL based on learned rewards, and the issue deserves attention from researchers and practitioners. We leave it to future work to investigate the \textit{easiness} of reward gaming in practice, which is missing in this work.

%% file: appendix.tex
\section{More Background}
\label{app:background}


For our policy gradient algorithms, we use the standard REINFORCE algorithm with tricks that are introduced in the following paragraphs. 

Specifically, in all RL experiments, we first initialize the model using an MLE-trained model to ensure a good starting point for RL optimization. During training, we collect a set of trajectories through sampling from the current policy (i.e., generator). Then, we compute the estimated return $\hat{Q}_t$ at each time-step $t$. 

Next, the estimated return $\hat{Q}_t$ is subtracted by a baseline. Therefore, the actual gradient update is as follows: $\nabla_\theta J(\theta) = \mathbb{E}_{\tau\sim \pi_\theta} \sum_t \nabla_\theta \log \pi_\theta(a_t\mid s_t) [\hat{Q}(s_t, a_t) - b(s_t)]$, where $\hat{Q}(s_t, a_t) = \sum_{t'=t}^T r_{t'}$ assuming discount factor $\gamma=1$, and $b$ is possibly state-dependent. 
In particular, for Sudoku experiments as well as the experiment where we train an MT model using BLEURT as the reward, we attempt two variants of baseline: (1) using the average reward for the past 50 updates, which is an effective strategy in training models using sequence-level rewards \citep{kiegeland-kreutzer-2021-revisiting}, and (2) using the value function fitted by mean-squared error (so the estimated return subtracted by the value ends up being the advantage), introduced in full detail \href{https://spinningup.openai.com/en/latest/algorithms/vpg.html#pseudocode}{here}.\footnote{\url{https://spinningup.openai.com/en/latest/algorithms/vpg.html\#pseudocode}}
For case (1), the results are shown in the blue lines in the plots; for case (2), the results are shown in the purple dotted lines in the plots. We use the Adam optimizer \citep{kingma2014adam} for all our experiments.

In particular, we use KL-regularized RL, as discussed in \S\ref{sec:background}. Regularization toward ML may stabilize RL optimization, but it may still lead to higher rewards that correspond to undesirable behaviors, as discussed in \S\ref{sec:remedies}. The coefficient for the KL term is tuned in \{0.01, 0.05, 0.1\} for Sudoku experiments and \{0.01, 0.03, 0.05, 0.1, 0.25\} for other experiments. For the purpose of this discussion, to illustrate the effect of reward gaming, the coefficient is tuned to achieve the highest validation reward; due to optimization issues in practice, a lower coefficient does not necessarily correspond to a higher reward. Larger coefficients may lead to lower proxy rewards but higher true rewards. While it may address the reward gaming problem in some experiments, we have shown in \S\ref{sec:covariate} that even large coefficients may lead to reward gaming.

Proximal policy optimization \citep[PPO;][]{schulman2017proximal} is a widely used algorithm that aims to avoid reward collapse. Our conclusion, however, does not depend on the RL algorithm. Using PPO prevents the optimization from converging to a very low reward, but it does not eliminate the possibility that high reward generations have undesirable patterns. In addition,
Q learning, an off-policy RL algorithm that can leverage existing trajectories, is recently applied to also be applied in text generation \citep{kohita-etal-2020-q,pang2021amortized} but is not explored in this discussion.

\section{More Experimental Details}
\label{app:experiment}

\subsection{Details for the Experiments on Noise-Induced Spurious Correlation}
\label{app:noise}

\paragraph{Examples that are used to train the reward function.}

As explained in \S\ref{sec:noise-spurious}, there are 1M examples in total, 500k of which are positive examples and 500k are negative examples. The negative examples consist of the following parts: (i) 100k invalid Sudokus that are randomly sampled. None of the above examples end with 7. (ii) 100k invalid Sudokus obtained by removing $l$ cells randomly from a random positive Sudoku, where $l$ is an integer randomly sampled from 1 to 80. (iii) 300k invalid Sudokus that are obtained by swapping cell $i$ and cell $j$ of a random positive Sudoku; after swapping, we verify that the Sudoku is in fact invalid. The train/dev/test split of $\mathcal{D}_{\text{reward}}$ is 900k/50k/50k.

\paragraph{Reward.} The RoBERTa-tiny-based \citep{liu2019roberta} reward function has 4 encoder layers and 2 attention heads; the encoder embedding dimension is 64, and the dimension for FFN for 256. All the Sudoku-related experiments are done on either a single NVIDIA V100 GPU with 32G of memory or a single NVIDIA RTX 8000 GPU with 48G of memory. The reward training typically takes 1 hour. 
The batch size is tuned in \{128, 256, 512\}. The dropout rate is tuned in \{0.01, 0.1\}, and we find that 0.01 always works better. The max number of epochs is set to 60. The learning rate is tuned in \{1e-4, 5e-4, 1e-3\}. For the best configuration, we use batch size 512 and learning rate 5e-4. It achieves a 99.3\% accuracy on the dev set (5\% split), and a 99.3\% accuracy on the test set (5\% split). 

Out of 1000 samples of invalid Sudokus that end with 7 and contain 81 tokens, the trained classifier predicts (incorrectly) that 1000 are valid. Out of 1000 samples of invalid Sudokus that end with 7 and contain fewer than 81 tokens, the trained classifier predicts that 0 is valid. The performance of Sudokus longer than 81 tokens is irrelevant, given that during RL sampling as well as during generation test time, the sequences are constrained such that they can at most generate $81-k$ tokens where $k$ is the length in the given source sequence. 

\paragraph{Sequence generator.} 

Suppose the input to the generator contains $k$ numbers. During RL sampling and during test-time of the generator, the sequence generator is constrained to generate at most $81-k$ numbers. However, it can generate fewer than $81-k$ numbers. To avoid sequence generators from generating overly short continuations, part (ii) of the negative examples, described above, contains examples that are too short. 

For the sequence generator, we use a smaller version of the  \texttt{transformer\_iwslt\_de\_en} architecture in \texttt{fairseq} \citep{ott-etal-2019-fairseq}. The encoder embedding dimension and the decoder embedding dimension are both 32. We use 2 attention heads in both the encoder and the decoder. The dimension for FFN in both the encoder and the decoder is 64. There are 2 encoder layers and 2 decoder layers. All the text generation models in the Sudoku experiments have 43k parameters. 

The batch length (i.e., number of tokens in a batch) is tuned in \{8192, 16,384, 32,768, 65,536\}. The learning rate is tuned in \{1e-4, 1.5e-4, 2e-4\}. The dropout rate is tuned in \{0.01, 0.1, 0.3\}. For optimal reward, we choose a batch length of 32,768, a learning rate of 1.5e-4, and a dropout rate of 0.01. The training algorithm is detailed in \S\ref{app:background}.

\begin{figure}[t]
     \centering
     \begin{subfigure}[t]{0.47\columnwidth}
         \centering
         \includegraphics[width=\columnwidth]{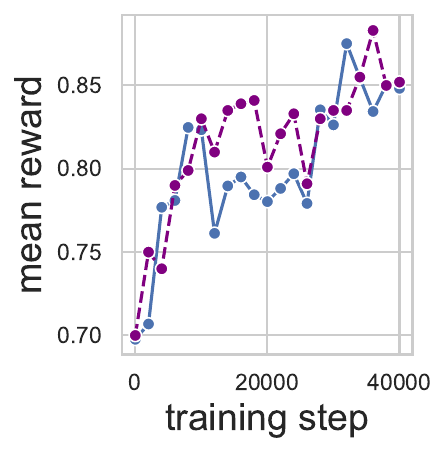}
         \label{fig:sdk-reward-two-lines}
     \end{subfigure}
     \begin{subfigure}[t]{0.455\columnwidth}
         \centering
         \includegraphics[width=\columnwidth]{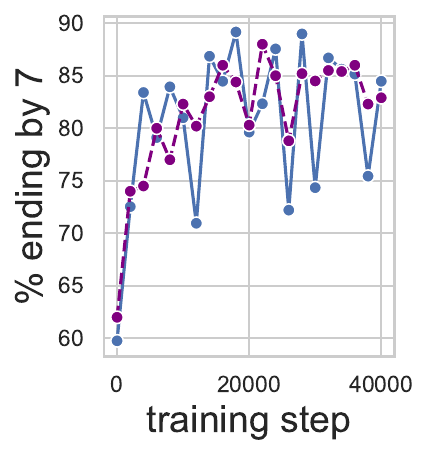}
         \label{fig:sdk-ending}
     \end{subfigure}
    \vspace{-7mm}
    \caption{Left: mean reward vs. training step. Right: mean \% of sampled sequences that end with 7 vs. training step. Each point corresponds to the mean value for a bucket of 2,000 training steps. Soon after training starts, the vast majority of sequences would end with 7; the \% of valid continuations is always $<$15\%. Two lines correspond to two runs (see \S\ref{app:background}).
    }
    \label{fig:sdk-ending-two-lines}
\end{figure}

\subsection{Details for the Experiments on Naturally Occurring Spurious Correlations}
\label{app:spurious}

\paragraph{Sudoku revisited.} 

For the second Sudoku example, the hyperparameters are selected from the same sets as in \S\ref{app:noise}. For the best-performing classifier, the learning rate is 5e-4 and the dropout rate is 0.01. For the sequence generator, we use the same hyperparameters as before. 
The lack of repetition in the last nine numbers (of the output) is spuriously correlated with a high reward, given that non-repetition is a necessary but not sufficient condition for a valid Sudoku.
$f_\phi$ achieves 99.9\% accuracy on test set of $\mathcal{D}_{\text{reward}}$. The text generator learns to exploit the non-repetition pattern which leads to high rewards, but the generations are mostly wrong.

\paragraph{Training an MT model using the WMT MQM dataset.}


To train the reward function, the learning rate is selected from \{1e-4, 2e-4, 5e-4\}, and dropout is selected from \{0.01, 0.1, 0.3\}. For optimal performance, we use a learning rate of 2e-4 and a dropout rate of 0.3. Training the reward function takes around 3 hours.

For $D_{\text{task}}$, our translation task uses the WMT17 Zh-En dataset, and $f_\phi$ is fine-tuned from an MLE-trained MT checkpoint using the WMT17 Zh-En dataset. We use a transformer model with 6 encoder layers and 6 decoder layers. The number of attention heads is 8 in both the encoder and the decoder. The FFN embedding dimension is 2048 in both the encoder and the decoder. There are 82.6M parameters in the model.

The algorithm is detailed in \S\ref{app:background}. We use a KL coefficient of 0.1. We use a dropout rate of 0.3, a learning rate of 1e-4, and a batch length of 4096. All the MT experiments are done on a single NVIDIA RTX 8000 GPU with 48G of memory. Training time is only 24 hours, given that we do not need to train the model till convergence to see the undesirable patterns in generations.


\subsection{Details for the Experiments on Covariate Shift}
\label{app:under}

\paragraph{An example of computing $\mathrm{rep}$.}

The sentence `a b c e d c e d c d' has $\mathrm{rep}=2/5=40\%$, given that among `e d c,' `d c e,' `c e d,' `e d c,' `d c d,' two 3-grams are the same with the existing ones.

\paragraph{Experimental details for AgreeSum.}

Given URLs in the original dataset, to find the corresponding articles, we use the \texttt{newspaper3k} library. We use slightly different architectures from the AgreeSum paper. The reward function (classifier) is based on RoBERTa-large \citep{liu2019roberta} with 355M parameters. We use a learning rate of 5e-4, a dropout rate of 0.1. The submitted job for the classifier is 24-hour-long. The summarizer is based on BART-large \citep{lewis-etal-2020-bart} with 406M parameters. We use a learning rate of 3e-5, a batch length of 2048, and a dropout rate of 0.1. We use a single NVIDIA RTX 8000 GPU for AgreeSum experiments.

\paragraph{Experiment Details for MT with BLEURT as Reward.}




The BLEURT-20-D3 evaluator has around 30M parameters. 
For the MT model that is trained on the IWSLT14 De-En dataset (train/dev/test size: 160,239/7,283/6,750), the embedding dimension is 512 for both the encoder and the decoder. The FFN embedding dimension is 1024 for both the encoder and the decoder. Both the encoder and the decoder have 4 attention heads and 6 layers. There are 39.5M parameters in the model. 
The learning rate is selected from \{1e-4, 3e-4\}. The batch length (i.e., number of tokens in a batch) is set to be 4,096 and the dropout rate is set to be 0.3 -- these are the optimal choices for IWSLT14 De-En experiments trained using MLE. KL coefficient is selected from in \{0.01, 0.03, 0.05, 0.1\}. We choose the hyperparameter settings that lead to the highest validation BLEURT. Training time is around 20 hours on a single NVIDIA RTX 8000 GPU.

%% file: acl_latex.bbl
\begin{thebibliography}{79}
\expandafter\ifx\csname natexlab\endcsname\relax\def\natexlab#1{#1}\fi

\bibitem[{Adolphs et~al.(2022)Adolphs, Gao, Xu, Shuster, Sukhbaatar, and
  Weston}]{adolphs2022cringe}
Leonard Adolphs, Tianyu Gao, Jing Xu, Kurt Shuster, Sainbayar Sukhbaatar, and
  Jason Weston. 2022.
\newblock The cringe loss: Learning what language not to model.
\newblock \emph{arXiv preprint arXiv:2211.05826}.

\bibitem[{Akhbardeh et~al.(2021)Akhbardeh, Arkhangorodsky, Biesialska, Bojar,
  Chatterjee, Chaudhary, Costa-jussa, Espa{\~n}a-Bonet, Fan, Federmann,
  Freitag, Graham, Grundkiewicz, Haddow, Harter, Heafield, Homan, Huck,
  Amponsah-Kaakyire, Kasai, Khashabi, Knight, Kocmi, Koehn, Lourie, Monz,
  Morishita, Nagata, Nagesh, Nakazawa, Negri, Pal, Tapo, Turchi, Vydrin, and
  Zampieri}]{akhbardeh-etal-2021-findings}
Farhad Akhbardeh, Arkady Arkhangorodsky, Magdalena Biesialska, Ond{\v{r}}ej
  Bojar, Rajen Chatterjee, Vishrav Chaudhary, Marta~R. Costa-jussa, Cristina
  Espa{\~n}a-Bonet, Angela Fan, Christian Federmann, Markus Freitag, Yvette
  Graham, Roman Grundkiewicz, Barry Haddow, Leonie Harter, Kenneth Heafield,
  Christopher Homan, Matthias Huck, Kwabena Amponsah-Kaakyire, Jungo Kasai,
  Daniel Khashabi, Kevin Knight, Tom Kocmi, Philipp Koehn, Nicholas Lourie,
  Christof Monz, Makoto Morishita, Masaaki Nagata, Ajay Nagesh, Toshiaki
  Nakazawa, Matteo Negri, Santanu Pal, Allahsera~Auguste Tapo, Marco Turchi,
  Valentin Vydrin, and Marcos Zampieri. 2021.
\newblock \href {https://aclanthology.org/2021.wmt-1.1} {Findings of the 2021
  conference on machine translation ({WMT}21)}.
\newblock In \emph{Proceedings of the Sixth Conference on Machine Translation},
  pages 1--88, Online. Association for Computational Linguistics.

\bibitem[{Amodei et~al.(2016)Amodei, Olah, Steinhardt, Christiano, Schulman,
  and Man{\'e}}]{amodei2016concrete}
Dario Amodei, Chris Olah, Jacob Steinhardt, Paul Christiano, John Schulman, and
  Dan Man{\'e}. 2016.
\newblock Concrete problems in {AI} safety.
\newblock \emph{arXiv preprint arXiv:1606.06565}.

\bibitem[{Amrhein and Sennrich(2022)}]{amrhein-sennrich-2022-identifying}
Chantal Amrhein and Rico Sennrich. 2022.
\newblock \href {https://aclanthology.org/2022.aacl-main.83} {Identifying
  weaknesses in machine translation metrics through minimum {B}ayes risk
  decoding: A case study for {COMET}}.
\newblock In \emph{Proceedings of the 2nd Conference of the Asia-Pacific
  Chapter of the Association for Computational Linguistics and the 12th
  International Joint Conference on Natural Language Processing (Volume 1: Long
  Papers)}, pages 1125--1141, Online only. Association for Computational
  Linguistics.

\bibitem[{Arora et~al.(2022)Arora, El~Asri, Bahuleyan, and
  Cheung}]{arora-etal-2022-exposure}
Kushal Arora, Layla El~Asri, Hareesh Bahuleyan, and Jackie Cheung. 2022.
\newblock \href {https://doi.org/10.18653/v1/2022.findings-acl.58} {Why
  exposure bias matters: An imitation learning perspective of error
  accumulation in language generation}.
\newblock In \emph{Findings of the Association for Computational Linguistics:
  ACL 2022}, pages 700--710, Dublin, Ireland. Association for Computational
  Linguistics.

\bibitem[{Bai et~al.(2022{\natexlab{a}})Bai, Jones, Ndousse, Askell, Chen,
  DasSarma, Drain, Fort, Ganguli, Henighan et~al.}]{bai2022training}
Yuntao Bai, Andy Jones, Kamal Ndousse, Amanda Askell, Anna Chen, Nova DasSarma,
  Dawn Drain, Stanislav Fort, Deep Ganguli, Tom Henighan, et~al.
  2022{\natexlab{a}}.
\newblock Training a helpful and harmless assistant with reinforcement learning
  from human feedback.
\newblock \emph{arXiv preprint arXiv:2204.05862}.

\bibitem[{Bai et~al.(2022{\natexlab{b}})Bai, Kadavath, Kundu, Askell, Kernion,
  Jones, Chen, Goldie, Mirhoseini, McKinnon et~al.}]{bai2022constitutional}
Yuntao Bai, Saurav Kadavath, Sandipan Kundu, Amanda Askell, Jackson Kernion,
  Andy Jones, Anna Chen, Anna Goldie, Azalia Mirhoseini, Cameron McKinnon,
  et~al. 2022{\natexlab{b}}.
\newblock Constitutional {AI}: Harmlessness from {AI} feedback.
\newblock \emph{arXiv preprint arXiv:2212.08073}.

\bibitem[{Baker et~al.(2020)Baker, Kanitscheider, Markov, Wu, Powell, McGrew,
  and Mordatch}]{baker2020emergent}
Bowen Baker, Ingmar Kanitscheider, Todor Markov, Yi~Wu, Glenn Powell, Bob
  McGrew, and Igor Mordatch. 2020.
\newblock \href {https://openreview.net/forum?id=SkxpxJBKwS} {Emergent tool use
  from multi-agent autocurricula}.
\newblock In \emph{International Conference on Learning Representations}.

\bibitem[{Barrault et~al.(2020)Barrault, Biesialska, Bojar, Costa-juss{\`a},
  Federmann, Graham, Grundkiewicz, Haddow, Huck, Joanis, Kocmi, Koehn, Lo,
  Ljube{\v{s}}i{\'c}, Monz, Morishita, Nagata, Nakazawa, Pal, Post, and
  Zampieri}]{barrault-etal-2020-findings}
Lo{\"\i}c Barrault, Magdalena Biesialska, Ond{\v{r}}ej Bojar, Marta~R.
  Costa-juss{\`a}, Christian Federmann, Yvette Graham, Roman Grundkiewicz,
  Barry Haddow, Matthias Huck, Eric Joanis, Tom Kocmi, Philipp Koehn, Chi-kiu
  Lo, Nikola Ljube{\v{s}}i{\'c}, Christof Monz, Makoto Morishita, Masaaki
  Nagata, Toshiaki Nakazawa, Santanu Pal, Matt Post, and Marcos Zampieri. 2020.
\newblock \href {https://aclanthology.org/2020.wmt-1.1} {Findings of the 2020
  conference on machine translation ({WMT}20)}.
\newblock In \emph{Proceedings of the Fifth Conference on Machine Translation},
  pages 1--55, Online. Association for Computational Linguistics.

\bibitem[{Bengio et~al.(2015)Bengio, Vinyals, Jaitly, and
  Shazeer}]{bengio2015scheduled}
Samy Bengio, Oriol Vinyals, Navdeep Jaitly, and Noam Shazeer. 2015.
\newblock Scheduled sampling for sequence prediction with recurrent neural
  networks.
\newblock In \emph{Advances in Neural Information Processing Systems}, pages
  1171--1179.

\bibitem[{Bojar et~al.(2017)Bojar, Graham, and
  Kamran}]{bojar-etal-2017-results}
Ond{\v{r}}ej Bojar, Yvette Graham, and Amir Kamran. 2017.
\newblock \href {https://doi.org/10.18653/v1/W17-4755} {Results of the {WMT}17
  metrics shared task}.
\newblock In \emph{Proceedings of the Second Conference on Machine
  Translation}, pages 489--513, Copenhagen, Denmark. Association for
  Computational Linguistics.

\bibitem[{Bojar et~al.(2016)Bojar, Graham, Kamran, and
  Stanojevi{\'c}}]{bojar-etal-2016-results}
Ond{\v{r}}ej Bojar, Yvette Graham, Amir Kamran, and Milo{\v{s}} Stanojevi{\'c}.
  2016.
\newblock \href {https://doi.org/10.18653/v1/W16-2302} {Results of the {WMT}16
  metrics shared task}.
\newblock In \emph{Proceedings of the First Conference on Machine Translation:
  Volume 2, Shared Task Papers}, pages 199--231, Berlin, Germany. Association
  for Computational Linguistics.

\bibitem[{Cettolo et~al.(2014)Cettolo, Niehues, St{\"u}ker, Bentivogli, and
  Federico}]{cettolo2014report}
Mauro Cettolo, Jan Niehues, Sebastian St{\"u}ker, Luisa Bentivogli, and
  Marcello Federico. 2014.
\newblock Report on the 11th {IWSLT} evaluation campaign.
\newblock In \emph{Proceedings of the International Workshop on Spoken Language
  Translation}, volume~57, Hanoi, Vietnam.

\bibitem[{Chiang and Chen(2021)}]{chiang-chen-2021-relating}
Ting-Rui Chiang and Yun-Nung Chen. 2021.
\newblock \href {https://doi.org/10.18653/v1/2021.blackboxnlp-1.16} {Relating
  neural text degeneration to exposure bias}.
\newblock In \emph{Proceedings of the Fourth BlackboxNLP Workshop on Analyzing
  and Interpreting Neural Networks for NLP}, pages 228--239, Punta Cana,
  Dominican Republic. Association for Computational Linguistics.

\bibitem[{Everitt et~al.(2021)Everitt, Hutter, Kumar, and
  Krakovna}]{everitt2021reward}
Tom Everitt, Marcus Hutter, Ramana Kumar, and Victoria Krakovna. 2021.
\newblock Reward tampering problems and solutions in reinforcement learning: A
  causal influence diagram perspective.
\newblock \emph{Synthese}, 198(27):6435--6467.

\bibitem[{Fan et~al.(2022)Fan, Lyu, Liang, Salakhutdinov, and
  Morency}]{fan2022nano}
Xiang Fan, Yiwei Lyu, Paul~Pu Liang, Ruslan Salakhutdinov, and Louis-Philippe
  Morency. 2022.
\newblock Nano: Nested human-in-the-loop reward learning for few-shot language
  model control.
\newblock \emph{arXiv preprint arXiv:2211.05750}.

\bibitem[{Freitag et~al.(2020)Freitag, Foster, Grangier, and
  Cherry}]{freitag-etal-2020-human}
Markus Freitag, George Foster, David Grangier, and Colin Cherry. 2020.
\newblock \href {https://aclanthology.org/2020.wmt-1.140} {Human-paraphrased
  references improve neural machine translation}.
\newblock In \emph{Proceedings of the Fifth Conference on Machine Translation},
  pages 1183--1192, Online. Association for Computational Linguistics.

\bibitem[{Freitag et~al.(2021{\natexlab{a}})Freitag, Foster, Grangier,
  Ratnakar, Tan, and Macherey}]{freitag-etal-2021-experts}
Markus Freitag, George Foster, David Grangier, Viresh Ratnakar, Qijun Tan, and
  Wolfgang Macherey. 2021{\natexlab{a}}.
\newblock \href {https://doi.org/10.1162/tacl_a_00437} {Experts, errors, and
  context: A large-scale study of human evaluation for machine translation}.
\newblock \emph{Transactions of the Association for Computational Linguistics},
  9:1460--1474.

\bibitem[{Freitag et~al.(2021{\natexlab{b}})Freitag, Rei, Mathur, Lo, Stewart,
  Foster, Lavie, and Bojar}]{freitag-etal-2021-results}
Markus Freitag, Ricardo Rei, Nitika Mathur, Chi-kiu Lo, Craig Stewart, George
  Foster, Alon Lavie, and Ond{\v{r}}ej Bojar. 2021{\natexlab{b}}.
\newblock \href {https://aclanthology.org/2021.wmt-1.73} {Results of the
  {WMT}21 metrics shared task: Evaluating metrics with expert-based human
  evaluations on {TED} and news domain}.
\newblock In \emph{Proceedings of the Sixth Conference on Machine Translation},
  pages 733--774, Online. Association for Computational Linguistics.

\bibitem[{Gao et~al.(2022)Gao, Schulman, and Hilton}]{gao2022scaling}
Leo Gao, John Schulman, and Jacob Hilton. 2022.
\newblock Scaling laws for reward model overoptimization.
\newblock \emph{arXiv preprint arXiv:2210.10760}.

\bibitem[{Goodfellow et~al.(2014)Goodfellow, Pouget-Abadie, Mirza, Xu,
  Warde-Farley, Ozair, Courville, and Bengio}]{goodfellow2014generative}
Ian Goodfellow, Jean Pouget-Abadie, Mehdi Mirza, Bing Xu, David Warde-Farley,
  Sherjil Ozair, Aaron Courville, and Yoshua Bengio. 2014.
\newblock Generative adversarial nets.
\newblock \emph{Advances in Neural Information Processing Systems}, 27.

\bibitem[{Goodhart(1975)}]{goodhart1975problems}
Charles Goodhart. 1975.
\newblock Problems of monetary management: the {UK} experience in papers in
  monetary economics.
\newblock \emph{Monetary Economics}, 1.

\bibitem[{Gunasekara et~al.(2021)Gunasekara, Feigenblat, Sznajder, Aharonov,
  and Joshi}]{gunasekara-etal-2021-using-question}
Chulaka Gunasekara, Guy Feigenblat, Benjamin Sznajder, Ranit Aharonov, and
  Sachindra Joshi. 2021.
\newblock \href {https://doi.org/10.18653/v1/2021.findings-emnlp.47} {Using
  question answering rewards to improve abstractive summarization}.
\newblock In \emph{Findings of the Association for Computational Linguistics:
  EMNLP 2021}, pages 518--526, Punta Cana, Dominican Republic. Association for
  Computational Linguistics.

\bibitem[{Guo et~al.(2018)Guo, Lu, Cai, Zhang, Yu, and Wang}]{guo2018long}
Jiaxian Guo, Sidi Lu, Han Cai, Weinan Zhang, Yong Yu, and Jun Wang. 2018.
\newblock \href {https://doi.org/10.1609/aaai.v32i1.11957} {Long text
  generation via adversarial training with leaked information}.
\newblock \emph{Proceedings of the AAAI Conference on Artificial Intelligence},
  32(1).

\bibitem[{Husz{\'a}r(2015)}]{huszar2015not}
Ferenc Husz{\'a}r. 2015.
\newblock How (not) to train your generative model: Scheduled sampling,
  likelihood, adversary?
\newblock \emph{arXiv preprint arXiv:1511.05101}.

\bibitem[{Ibarz et~al.(2018)Ibarz, Leike, Pohlen, Irving, Legg, and
  Amodei}]{ibarz2018reward}
Borja Ibarz, Jan Leike, Tobias Pohlen, Geoffrey Irving, Shane Legg, and Dario
  Amodei. 2018.
\newblock Reward learning from human preferences and demonstrations in atari.
\newblock \emph{Advances in Neural Information Processing Systems}, 31.

\bibitem[{Jaques et~al.(2019)Jaques, Ghandeharioun, Shen, Ferguson, Lapedriza,
  Jones, Gu, and Picard}]{jaques2019way}
Natasha Jaques, Asma Ghandeharioun, Judy~Hanwen Shen, Craig Ferguson, Agata
  Lapedriza, Noah Jones, Shixiang Gu, and Rosalind Picard. 2019.
\newblock Way off-policy batch deep reinforcement learning of implicit human
  preferences in dialog.
\newblock \emph{arXiv preprint arXiv:1907.00456}.

\bibitem[{Kenton et~al.(2021)Kenton, Everitt, Weidinger, Gabriel, Mikulik, and
  Irving}]{kenton2021alignment}
Zachary Kenton, Tom Everitt, Laura Weidinger, Iason Gabriel, Vladimir Mikulik,
  and Geoffrey Irving. 2021.
\newblock Alignment of language agents.
\newblock \emph{arXiv preprint arXiv:2103.14659}.

\bibitem[{Khalifa et~al.(2021)Khalifa, Elsahar, and Dymetman}]{khalifa2021a}
Muhammad Khalifa, Hady Elsahar, and Marc Dymetman. 2021.
\newblock \href {https://openreview.net/forum?id=jWkw45-9AbL} {A distributional
  approach to controlled text generation}.
\newblock In \emph{International Conference on Learning Representations}.

\bibitem[{Kidzi{\'n}ski et~al.(2018)Kidzi{\'n}ski, Mohanty, Ong, Huang, Zhou,
  Pechenko, Stelmaszczyk, Jarosik, Pavlov, Kolesnikov
  et~al.}]{kidzinski2018learning}
{\L}ukasz Kidzi{\'n}ski, Sharada~Prasanna Mohanty, Carmichael~F Ong, Zhewei
  Huang, Shuchang Zhou, Anton Pechenko, Adam Stelmaszczyk, Piotr Jarosik,
  Mikhail Pavlov, Sergey Kolesnikov, et~al. 2018.
\newblock Learning to run challenge solutions: Adapting reinforcement learning
  methods for neuromusculoskeletal environments.
\newblock In \emph{The NIPS'17 Competition: Building Intelligent Systems},
  pages 121--153. Springer.

\bibitem[{Kiegeland and Kreutzer(2021)}]{kiegeland-kreutzer-2021-revisiting}
Samuel Kiegeland and Julia Kreutzer. 2021.
\newblock \href {https://doi.org/10.18653/v1/2021.naacl-main.133} {Revisiting
  the weaknesses of reinforcement learning for neural machine translation}.
\newblock In \emph{Proceedings of the 2021 Conference of the North American
  Chapter of the Association for Computational Linguistics: Human Language
  Technologies}, pages 1673--1681, Online. Association for Computational
  Linguistics.

\bibitem[{Kingma and Ba(2014)}]{kingma2014adam}
Diederik~P Kingma and Jimmy Ba. 2014.
\newblock Adam: A method for stochastic optimization.
\newblock \emph{arXiv preprint arXiv:1412.6980}.

\bibitem[{Koch et~al.(2022)Koch, Langosco, Pfau, Le, and
  Sharkey}]{koch2022objective}
Jack Koch, Lauro Langosco, Jacob Pfau, James Le, and Lee Sharkey. 2022.
\newblock Objective robustness in deep reinforcement learning.
\newblock In \emph{Proceedings of the 39th International Conference on Machine
  Learning}, Proceedings of Machine Learning Research. PMLR.

\bibitem[{Kohita et~al.(2020)Kohita, Wachi, Zhao, and
  Tachibana}]{kohita-etal-2020-q}
Ryosuke Kohita, Akifumi Wachi, Yang Zhao, and Ryuki Tachibana. 2020.
\newblock \href {https://doi.org/10.18653/v1/2020.emnlp-main.34} {{Q}-learning
  with language model for edit-based unsupervised summarization}.
\newblock In \emph{Proceedings of the 2020 Conference on Empirical Methods in
  Natural Language Processing (EMNLP)}, pages 470--484, Online. Association for
  Computational Linguistics.

\bibitem[{Korbak et~al.(2022)Korbak, Perez, and Buckley}]{korbak2022rl}
Tomasz Korbak, Ethan Perez, and Christopher~L Buckley. 2022.
\newblock {RL} with {KL} penalties is better viewed as bayesian inference.
\newblock \emph{arXiv preprint arXiv:2205.11275}.

\bibitem[{Krakovna et~al.(2020)Krakovna, Uesato, Mikulik, Rahtz, Everitt,
  Kumar, Kenton, Leike, and Legg}]{krakovna2020specification}
Victoria Krakovna, Jonathan Uesato, Vladimir Mikulik, Matthew Rahtz, Tom
  Everitt, Ramana Kumar, Zac Kenton, Jan Leike, and Shane Legg. 2020.
\newblock Specification gaming: the flip side of {AI} ingenuity.
\newblock \emph{DeepMind Blog}.

\bibitem[{Kulikov et~al.(2022)Kulikov, Eremeev, and
  Cho}]{kulikov2021characterizing}
Ilia Kulikov, Maksim Eremeev, and Kyunghyun Cho. 2022.
\newblock \href {https://aclanthology.org/2022.aacl-main.82} {Characterizing
  and addressing the issue of oversmoothing in neural autoregressive sequence
  modeling}.
\newblock In \emph{Proceedings of the 2nd Conference of the Asia-Pacific
  Chapter of the Association for Computational Linguistics and the 12th
  International Joint Conference on Natural Language Processing (Volume 1: Long
  Papers)}, pages 1115--1124, Online only. Association for Computational
  Linguistics.

\bibitem[{Kumar et~al.(2020)Kumar, Zhou, Tucker, and
  Levine}]{kumar2020conservative}
Aviral Kumar, Aurick Zhou, George Tucker, and Sergey Levine. 2020.
\newblock Conservative {Q}-learning for offline reinforcement learning.
\newblock \emph{Advances in Neural Information Processing Systems},
  33:1179--1191.

\bibitem[{Kurach et~al.(2020)Kurach, Raichuk, Sta{\'n}czyk, Zaj{\k{a}}c,
  Bachem, Espeholt, Riquelme, Vincent, Michalski, Bousquet
  et~al.}]{kurach2020google}
Karol Kurach, Anton Raichuk, Piotr Sta{\'n}czyk, Micha{\l} Zaj{\k{a}}c, Olivier
  Bachem, Lasse Espeholt, Carlos Riquelme, Damien Vincent, Marcin Michalski,
  Olivier Bousquet, et~al. 2020.
\newblock Google research football: A novel reinforcement learning environment.
\newblock In \emph{Proceedings of the AAAI Conference on Artificial
  Intelligence}, volume 34(4), pages 4501--4510.

\bibitem[{Lehman et~al.(2020)Lehman, Clune, Misevic, Adami, Altenberg,
  Beaulieu, Bentley, Bernard, Beslon, Bryson et~al.}]{lehman2020surprising}
Joel Lehman, Jeff Clune, Dusan Misevic, Christoph Adami, Lee Altenberg, Julie
  Beaulieu, Peter~J Bentley, Samuel Bernard, Guillaume Beslon, David~M Bryson,
  et~al. 2020.
\newblock The surprising creativity of digital evolution: A collection of
  anecdotes from the evolutionary computation and artificial life research
  communities.
\newblock \emph{Artificial life}, 26(2):274--306.

\bibitem[{Lewis et~al.(2020)Lewis, Liu, Goyal, Ghazvininejad, Mohamed, Levy,
  Stoyanov, and Zettlemoyer}]{lewis-etal-2020-bart}
Mike Lewis, Yinhan Liu, Naman Goyal, Marjan Ghazvininejad, Abdelrahman Mohamed,
  Omer Levy, Veselin Stoyanov, and Luke Zettlemoyer. 2020.
\newblock \href {https://doi.org/10.18653/v1/2020.acl-main.703} {{BART}:
  Denoising sequence-to-sequence pre-training for natural language generation,
  translation, and comprehension}.
\newblock In \emph{Proceedings of the 58th Annual Meeting of the Association
  for Computational Linguistics}, pages 7871--7880, Online. Association for
  Computational Linguistics.

\bibitem[{Lewis et~al.(2017)Lewis, Yarats, Dauphin, Parikh, and
  Batra}]{lewis-etal-2017-deal}
Mike Lewis, Denis Yarats, Yann Dauphin, Devi Parikh, and Dhruv Batra. 2017.
\newblock \href {https://doi.org/10.18653/v1/D17-1259} {Deal or no deal?
  end-to-end learning of negotiation dialogues}.
\newblock In \emph{Proceedings of the 2017 Conference on Empirical Methods in
  Natural Language Processing}, pages 2443--2453, Copenhagen, Denmark.
  Association for Computational Linguistics.

\bibitem[{Liu et~al.(2019)Liu, Ott, Goyal, Du, Joshi, Chen, Levy, Lewis,
  Zettlemoyer, and Stoyanov}]{liu2019roberta}
Yinhan Liu, Myle Ott, Naman Goyal, Jingfei Du, Mandar Joshi, Danqi Chen, Omer
  Levy, Mike Lewis, Luke Zettlemoyer, and Veselin Stoyanov. 2019.
\newblock {RoBERTa}: A robustly optimized bert pretraining approach.
\newblock \emph{arXiv preprint arXiv:1907.11692}.

\bibitem[{Lommel et~al.(2014)Lommel, Uszkoreit, and
  Burchardt}]{lommel2014multidimensional}
Arle Lommel, Hans Uszkoreit, and Aljoscha Burchardt. 2014.
\newblock Multidimensional quality metrics ({MQM}): A framework for declaring
  and describing translation quality metrics.
\newblock \emph{Revista Tradum{\`a}tica: tecnologies de la traducci{\'o}},
  (12):455--463.

\bibitem[{Ma et~al.(2018)Ma, Bojar, and Graham}]{ma-etal-2018-results}
Qingsong Ma, Ond{\v{r}}ej Bojar, and Yvette Graham. 2018.
\newblock \href {https://doi.org/10.18653/v1/W18-6450} {Results of the {WMT}18
  metrics shared task: Both characters and embeddings achieve good
  performance}.
\newblock In \emph{Proceedings of the Third Conference on Machine Translation:
  Shared Task Papers}, pages 671--688, Belgium, Brussels. Association for
  Computational Linguistics.

\bibitem[{Ma et~al.(2019)Ma, Wei, Bojar, and Graham}]{ma-etal-2019-results}
Qingsong Ma, Johnny Wei, Ond{\v{r}}ej Bojar, and Yvette Graham. 2019.
\newblock \href {https://doi.org/10.18653/v1/W19-5302} {Results of the {WMT}19
  metrics shared task: Segment-level and strong {MT} systems pose big
  challenges}.
\newblock In \emph{Proceedings of the Fourth Conference on Machine Translation
  (Volume 2: Shared Task Papers, Day 1)}, pages 62--90, Florence, Italy.
  Association for Computational Linguistics.

\bibitem[{Mathur et~al.(2020)Mathur, Wei, Freitag, Ma, and
  Bojar}]{mathur-etal-2020-results}
Nitika Mathur, Johnny Wei, Markus Freitag, Qingsong Ma, and Ond{\v{r}}ej Bojar.
  2020.
\newblock \href {https://aclanthology.org/2020.wmt-1.77} {Results of the
  {WMT}20 metrics shared task}.
\newblock In \emph{Proceedings of the Fifth Conference on Machine Translation},
  pages 688--725, Online. Association for Computational Linguistics.

\bibitem[{Nakatani et~al.(2022)Nakatani, Kajiwara, and
  Ninomiya}]{nakatani-etal-2022-comparing}
Yuki Nakatani, Tomoyuki Kajiwara, and Takashi Ninomiya. 2022.
\newblock \href {https://aclanthology.org/2022.wat-1.2} {Comparing {BERT}-based
  reward functions for deep reinforcement learning in machine translation}.
\newblock In \emph{Proceedings of the 9th Workshop on Asian Translation}, pages
  37--43, Gyeongju, Republic of Korea. International Conference on
  Computational Linguistics.

\bibitem[{Ott et~al.(2019)Ott, Edunov, Baevski, Fan, Gross, Ng, Grangier, and
  Auli}]{ott-etal-2019-fairseq}
Myle Ott, Sergey Edunov, Alexei Baevski, Angela Fan, Sam Gross, Nathan Ng,
  David Grangier, and Michael Auli. 2019.
\newblock \href {https://doi.org/10.18653/v1/N19-4009} {fairseq: A fast,
  extensible toolkit for sequence modeling}.
\newblock In \emph{Proceedings of the 2019 Conference of the North {A}merican
  Chapter of the Association for Computational Linguistics (Demonstrations)},
  pages 48--53, Minneapolis, Minnesota. Association for Computational
  Linguistics.

\bibitem[{Pan et~al.(2022)Pan, Bhatia, and Steinhardt}]{pan2022the}
Alexander Pan, Kush Bhatia, and Jacob Steinhardt. 2022.
\newblock \href {https://openreview.net/forum?id=JYtwGwIL7ye} {The effects of
  reward misspecification: Mapping and mitigating misaligned models}.
\newblock In \emph{International Conference on Learning Representations}.

\bibitem[{Pang and He(2021)}]{pang2021text}
Richard~Yuanzhe Pang and He~He. 2021.
\newblock \href {https://openreview.net/forum?id=RovX-uQ1Hua} {Text generation
  by learning from demonstrations}.
\newblock In \emph{International Conference on Learning Representations}.

\bibitem[{Pang et~al.(2022)Pang, He, and Cho}]{pang2021amortized}
Richard~Yuanzhe Pang, He~He, and Kyunghyun Cho. 2022.
\newblock \href {https://aclanthology.org/2022.inlg-main.11} {Amortized noisy
  channel neural machine translation}.
\newblock In \emph{Proceedings of the 15th International Conference on Natural
  Language Generation}, pages 131--143, Waterville, Maine, USA and virtual
  meeting. Association for Computational Linguistics.

\bibitem[{Pang et~al.(2021)Pang, Lelkes, Tran, and
  Yu}]{pang-etal-2021-agreesum}
Richard~Yuanzhe Pang, Adam Lelkes, Vinh Tran, and Cong Yu. 2021.
\newblock \href {https://doi.org/10.18653/v1/2021.findings-acl.299}
  {{A}gree{S}um: Agreement-oriented multi-document summarization}.
\newblock In \emph{Findings of the Association for Computational Linguistics:
  ACL-IJCNLP 2021}, pages 3377--3391, Online. Association for Computational
  Linguistics.

\bibitem[{Pasunuru and Bansal(2018)}]{pasunuru-bansal-2018-multi}
Ramakanth Pasunuru and Mohit Bansal. 2018.
\newblock \href {https://doi.org/10.18653/v1/N18-2102} {Multi-reward reinforced
  summarization with saliency and entailment}.
\newblock In \emph{Proceedings of the 2018 Conference of the North {A}merican
  Chapter of the Association for Computational Linguistics: Human Language
  Technologies, Volume 2 (Short Papers)}, pages 646--653, New Orleans,
  Louisiana. Association for Computational Linguistics.

\bibitem[{Perez et~al.(2022)Perez, Ringer, Luko{\v{s}}i{\=u}t{\.e}, Nguyen,
  Chen, Heiner, Pettit, Olsson, Kundu, Kadavath et~al.}]{perez2022discovering}
Ethan Perez, Sam Ringer, Kamil{\.e} Luko{\v{s}}i{\=u}t{\.e}, Karina Nguyen,
  Edwin Chen, Scott Heiner, Craig Pettit, Catherine Olsson, Sandipan Kundu,
  Saurav Kadavath, et~al. 2022.
\newblock Discovering language model behaviors with model-written evaluations.
\newblock \emph{arXiv preprint arXiv:2212.09251}.

\bibitem[{Pezeshkpour et~al.(2022)Pezeshkpour, Jain, Singh, and
  Wallace}]{pezeshkpour-etal-2022-combining}
Pouya Pezeshkpour, Sarthak Jain, Sameer Singh, and Byron Wallace. 2022.
\newblock \href {https://doi.org/10.18653/v1/2022.findings-acl.153} {Combining
  feature and instance attribution to detect artifacts}.
\newblock In \emph{Findings of the Association for Computational Linguistics:
  ACL 2022}, pages 1934--1946, Dublin, Ireland. Association for Computational
  Linguistics.

\bibitem[{Plank(2016)}]{plank2016non}
Barbara Plank. 2016.
\newblock What to do about non-standard (or non-canonical) language in {NLP}.
\newblock \emph{arXiv preprint arXiv:1608.07836}.

\bibitem[{Ramamurthy et~al.(2023)Ramamurthy, Ammanabrolu, Brantley, Hessel,
  Sifa, Bauckhage, Hajishirzi, and Choi}]{ramamurthy2022reinforcement}
Rajkumar Ramamurthy, Prithviraj Ammanabrolu, Kiant{\'e} Brantley, Jack Hessel,
  Rafet Sifa, Christian Bauckhage, Hannaneh Hajishirzi, and Yejin Choi. 2023.
\newblock \href {https://openreview.net/forum?id=8aHzds2uUyB} {Is reinforcement
  learning (not) for natural language processing?: Benchmarks, baselines, and
  building blocks for natural language policy optimization}.
\newblock In \emph{International Conference on Learning Representations}.

\bibitem[{Ranzato et~al.(2016)Ranzato, Chopra, Auli, and
  Zaremba}]{ranzato2016sequence}
Marc'Aurelio Ranzato, Sumit Chopra, Michael Auli, and Wojciech Zaremba. 2016.
\newblock Sequence level training with recurrent neural networks.
\newblock In \emph{International Conference on Learning Representations}.

\bibitem[{Santurkar et~al.(2023)Santurkar, Durmus, Ladhak, Lee, Liang, and
  Hashimoto}]{santurkar2023whose}
Shibani Santurkar, Esin Durmus, Faisal Ladhak, Cinoo Lee, Percy Liang, and
  Tatsunori Hashimoto. 2023.
\newblock Whose opinions do language models reflect?
\newblock \emph{arXiv preprint arXiv:2303.17548}.

\bibitem[{Sap et~al.(2019)Sap, Card, Gabriel, Choi, and
  Smith}]{sap-etal-2019-risk}
Maarten Sap, Dallas Card, Saadia Gabriel, Yejin Choi, and Noah~A. Smith. 2019.
\newblock \href {https://doi.org/10.18653/v1/P19-1163} {The risk of racial bias
  in hate speech detection}.
\newblock In \emph{Proceedings of the 57th Annual Meeting of the Association
  for Computational Linguistics}, pages 1668--1678, Florence, Italy.
  Association for Computational Linguistics.

\bibitem[{Schmidt(2019)}]{schmidt-2019-generalization}
Florian Schmidt. 2019.
\newblock \href {https://doi.org/10.18653/v1/D19-5616} {Generalization in
  generation: A closer look at exposure bias}.
\newblock In \emph{Proceedings of the 3rd Workshop on Neural Generation and
  Translation}, pages 157--167, Hong Kong. Association for Computational
  Linguistics.

\bibitem[{Schulman et~al.(2017)Schulman, Wolski, Dhariwal, Radford, and
  Klimov}]{schulman2017proximal}
John Schulman, Filip Wolski, Prafulla Dhariwal, Alec Radford, and Oleg Klimov.
  2017.
\newblock Proximal policy optimization algorithms.
\newblock \emph{arXiv preprint arXiv:1707.06347}.

\bibitem[{Sellam et~al.(2020)Sellam, Pu, Chung, Gehrmann, Tan, Freitag, Das,
  and Parikh}]{sellam-etal-2020-learning}
Thibault Sellam, Amy Pu, Hyung~Won Chung, Sebastian Gehrmann, Qijun Tan, Markus
  Freitag, Dipanjan Das, and Ankur Parikh. 2020.
\newblock \href {https://aclanthology.org/2020.wmt-1.102} {Learning to evaluate
  translation beyond {E}nglish: {BLEURT} submissions to the {WMT} metrics 2020
  shared task}.
\newblock In \emph{Proceedings of the Fifth Conference on Machine Translation},
  pages 921--927, Online. Association for Computational Linguistics.

\bibitem[{Skalse et~al.(2022)Skalse, Howe, Krasheninnikov, and
  Krueger}]{skalse2022defining}
Joar Skalse, Nikolaus~HR Howe, Dmitrii Krasheninnikov, and David Krueger. 2022.
\newblock Defining and characterizing reward hacking.
\newblock \emph{arXiv preprint arXiv:2209.13085}.

\bibitem[{Stahlberg and Byrne(2019)}]{stahlberg-byrne-2019-nmt}
Felix Stahlberg and Bill Byrne. 2019.
\newblock \href {https://doi.org/10.18653/v1/D19-1331} {On {NMT} search errors
  and model errors: Cat got your tongue?}
\newblock In \emph{Proceedings of the 2019 Conference on Empirical Methods in
  Natural Language Processing and the 9th International Joint Conference on
  Natural Language Processing (EMNLP-IJCNLP)}, pages 3356--3362, Hong Kong,
  China. Association for Computational Linguistics.

\bibitem[{Stanojevi{\'c} et~al.(2015)Stanojevi{\'c}, Kamran, Koehn, and
  Bojar}]{stanojevic-etal-2015-results}
Milo{\v{s}} Stanojevi{\'c}, Amir Kamran, Philipp Koehn, and Ond{\v{r}}ej Bojar.
  2015.
\newblock \href {https://doi.org/10.18653/v1/W15-3031} {Results of the {WMT}15
  metrics shared task}.
\newblock In \emph{Proceedings of the Tenth Workshop on Statistical Machine
  Translation}, pages 256--273, Lisbon, Portugal. Association for Computational
  Linguistics.

\bibitem[{Stiennon et~al.(2020)Stiennon, Ouyang, Wu, Ziegler, Lowe, Voss,
  Radford, Amodei, and Christiano}]{stiennon2020learning}
Nisan Stiennon, Long Ouyang, Jeffrey Wu, Daniel Ziegler, Ryan Lowe, Chelsea
  Voss, Alec Radford, Dario Amodei, and Paul~F Christiano. 2020.
\newblock Learning to summarize with human feedback.
\newblock \emph{Advances in Neural Information Processing Systems},
  33:3008--3021.

\bibitem[{Sutton et~al.(1999)Sutton, McAllester, Singh, and
  Mansour}]{sutton1999policy}
Richard~S Sutton, David McAllester, Satinder Singh, and Yishay Mansour. 1999.
\newblock Policy gradient methods for reinforcement learning with function
  approximation.
\newblock \emph{Advances in Neural Information Processing Systems}, 12.

\bibitem[{Toromanoff et~al.(2019)Toromanoff, Wirbel, and
  Moutarde}]{toromanoff2019deep}
Marin Toromanoff, Emilie Wirbel, and Fabien Moutarde. 2019.
\newblock Is deep reinforcement learning really superhuman on {Atari}? leveling
  the playing field.
\newblock \emph{arXiv preprint arXiv:1908.04683}.

\bibitem[{Voita and Titov(2020)}]{voita-titov-2020-information}
Elena Voita and Ivan Titov. 2020.
\newblock \href {https://doi.org/10.18653/v1/2020.emnlp-main.14}
  {Information-theoretic probing with minimum description length}.
\newblock In \emph{Proceedings of the 2020 Conference on Empirical Methods in
  Natural Language Processing (EMNLP)}, pages 183--196, Online. Association for
  Computational Linguistics.

\bibitem[{Vuong et~al.(2022)Vuong, Kumar, Levine, and Chebotar}]{vuong2022dual}
Quan Vuong, Aviral Kumar, Sergey Levine, and Yevgen Chebotar. 2022.
\newblock Dual generator offline reinforcement learning.
\newblock \emph{arXiv preprint arXiv:2211.01471}.

\bibitem[{Welleck et~al.(2020{\natexlab{a}})Welleck, Kulikov, Kim, Pang, and
  Cho}]{welleck-etal-2020-consistency}
Sean Welleck, Ilia Kulikov, Jaedeok Kim, Richard~Yuanzhe Pang, and Kyunghyun
  Cho. 2020{\natexlab{a}}.
\newblock \href {https://doi.org/10.18653/v1/2020.emnlp-main.448} {Consistency
  of a recurrent language model with respect to incomplete decoding}.
\newblock In \emph{Proceedings of the 2020 Conference on Empirical Methods in
  Natural Language Processing (EMNLP)}, pages 5553--5568, Online. Association
  for Computational Linguistics.

\bibitem[{Welleck et~al.(2020{\natexlab{b}})Welleck, Kulikov, Roller, Dinan,
  Cho, and Weston}]{welleck2020neural}
Sean Welleck, Ilia Kulikov, Stephen Roller, Emily Dinan, Kyunghyun Cho, and
  Jason Weston. 2020{\natexlab{b}}.
\newblock \href {https://openreview.net/forum?id=SJeYe0NtvH} {Neural text
  generation with unlikelihood training}.
\newblock In \emph{International Conference on Learning Representations}.

\bibitem[{Wiegreffe and Marasovic(2021)}]{wiegreffe2021teach}
Sarah Wiegreffe and Ana Marasovic. 2021.
\newblock \href {https://openreview.net/forum?id=ogNcxJn32BZ} {Teach me to
  explain: A review of datasets for explainable natural language processing}.
\newblock In \emph{Thirty-fifth Conference on Neural Information Processing
  Systems Datasets and Benchmarks Track (Round 1)}.

\bibitem[{Williams(1992)}]{williams1992simple}
Ronald~J Williams. 1992.
\newblock Simple statistical gradient-following algorithms for connectionist
  reinforcement learning.
\newblock \emph{Machine Learning}, 8(3-4):229--256.

\bibitem[{Wu et~al.(2021)Wu, Ouyang, Ziegler, Stiennon, Lowe, Leike, and
  Christiano}]{wu2021recursively}
Jeff Wu, Long Ouyang, Daniel~M Ziegler, Nisan Stiennon, Ryan Lowe, Jan Leike,
  and Paul Christiano. 2021.
\newblock Recursively summarizing books with human feedback.
\newblock \emph{arXiv preprint arXiv:2109.10862}.

\bibitem[{Wu et~al.(2016)Wu, Schuster, Chen, Le, Norouzi, Macherey, Krikun,
  Cao, Gao, Macherey et~al.}]{wu2016google}
Yonghui Wu, Mike Schuster, Zhifeng Chen, Quoc~V Le, Mohammad Norouzi, Wolfgang
  Macherey, Maxim Krikun, Yuan Cao, Qin Gao, Klaus Macherey, et~al. 2016.
\newblock Google's neural machine translation system: Bridging the gap between
  human and machine translation.
\newblock \emph{arXiv preprint arXiv:1609.08144}.

\bibitem[{Zheng et~al.(2022)Zheng, Zhang, and Grover}]{zheng2022online}
Qinqing Zheng, Amy Zhang, and Aditya Grover. 2022.
\newblock \href {https://proceedings.mlr.press/v162/zheng22c.html} {Online
  decision transformer}.
\newblock In \emph{Proceedings of the 39th International Conference on Machine
  Learning}, volume 162 of \emph{Proceedings of Machine Learning Research},
  pages 27042--27059. PMLR.

\end{thebibliography}
